\documentclass[10pt,twocolumn,letterpaper]{article}

\usepackage{iccv}              % To produce the CAMERA-READY version
\definecolor{iccvblue}{rgb}{0.21,0.49,0.74}
\usepackage[pagebackref,breaklinks,colorlinks,allcolors=iccvblue]{hyperref}
\usepackage{enumitem}
\usepackage{graphicx}
\usepackage[linesnumbered,ruled,vlined]{algorithm2e}
\usepackage{booktabs}
\usepackage{multirow}
\usepackage{tabularx}
\usepackage{array}
\usepackage{pgfplots}
\pgfplotsset{compat=1.17}

\def\paperID{*****} % *** Enter the Paper ID here
\def\confName{ICCV}
\def\confYear{2025}

\title{Pi-GPS: Enhancing Geometry Problem Solving by Unleashing the Power of Diagrammatic Information}

\author{
 Junbo Zhao$^1$\textsuperscript{*}
 \quad Ting Zhang$^1$\textsuperscript{*}
 \quad Jiayu Sun$^1$
 \quad Mi Tian$^2$
 \quad Hua Huang$^1$\textsuperscript{†} \\
 {$^1$Beijing Normal University \quad $^2$TAL}\\
 {}
}

\begin{document}
\maketitle

{
  \renewcommand{\thefootnote}%
    {\fnsymbol{footnote}}
  \footnotetext[1]{Equal contribution.}
  \footnotetext[2]{Corresponding author.}
}

\begin{abstract}
Geometry problem solving has garnered increasing attention due to its potential applications in intelligent education field. Inspired by the observation that text often introduces ambiguities that diagrams can clarify, this paper presents Pi-GPS, a novel framework that unleashes the power of diagrammatic information to resolve textual ambiguities, an aspect largely overlooked in prior research. Specifically, we design a micro module comprising a rectifier and verifier: the rectifier employs MLLMs to disambiguate text based on the diagrammatic context, while the verifier ensures the rectified output adherence to geometric rules, mitigating model hallucinations.
Additionally, we explore the impact of LLMs in theorem predictor based on the disambiguated formal language. Empirical results demonstrate that Pi-GPS surpasses state-of-the-art models, achieving a nearly 10\% improvement on Geometry3K over prior neural-symbolic approaches. We hope this work highlights the significance of resolving textual ambiguity in multimodal mathematical reasoning, a crucial factor limiting performance.
\end{abstract}

\section{Introduction}
Geometry Problem Solving (GPS) aims to derive solutions from a textual problem description and its corresponding diagram. As a distinct and pivotal aspect of multimodal mathematical reasoning, GPS requires a nuanced understanding of visual shapes, intricate spatial relationships, symbolic abstraction, and logical inference across both textual and diagrammatic inputs. This makes it a long-standing challenge in mathematical reasoning and artificial intelligence~~\cite{gelernter1960empirical,wen1986basic, sachan2017learning, GeoQA,lu2021inter}. While recent notable milestones~~\cite{alphaproof,AlphaGeo} such as the gold-medal-level solution to geometry problems using AlphaGeometry2~~\cite{AlphaGeometry2} have exhibited remarkable achievements, these efforts predominantly focus on language processing, neglecting the diagrammatic component of the problem. However, GPS transcends language-based reasoning, demanding a profound understanding and manipulation of diagrammatic information, an enduring challenge in the field.
\begin{figure}[t]
  \centering
  \includegraphics[width=\columnwidth]{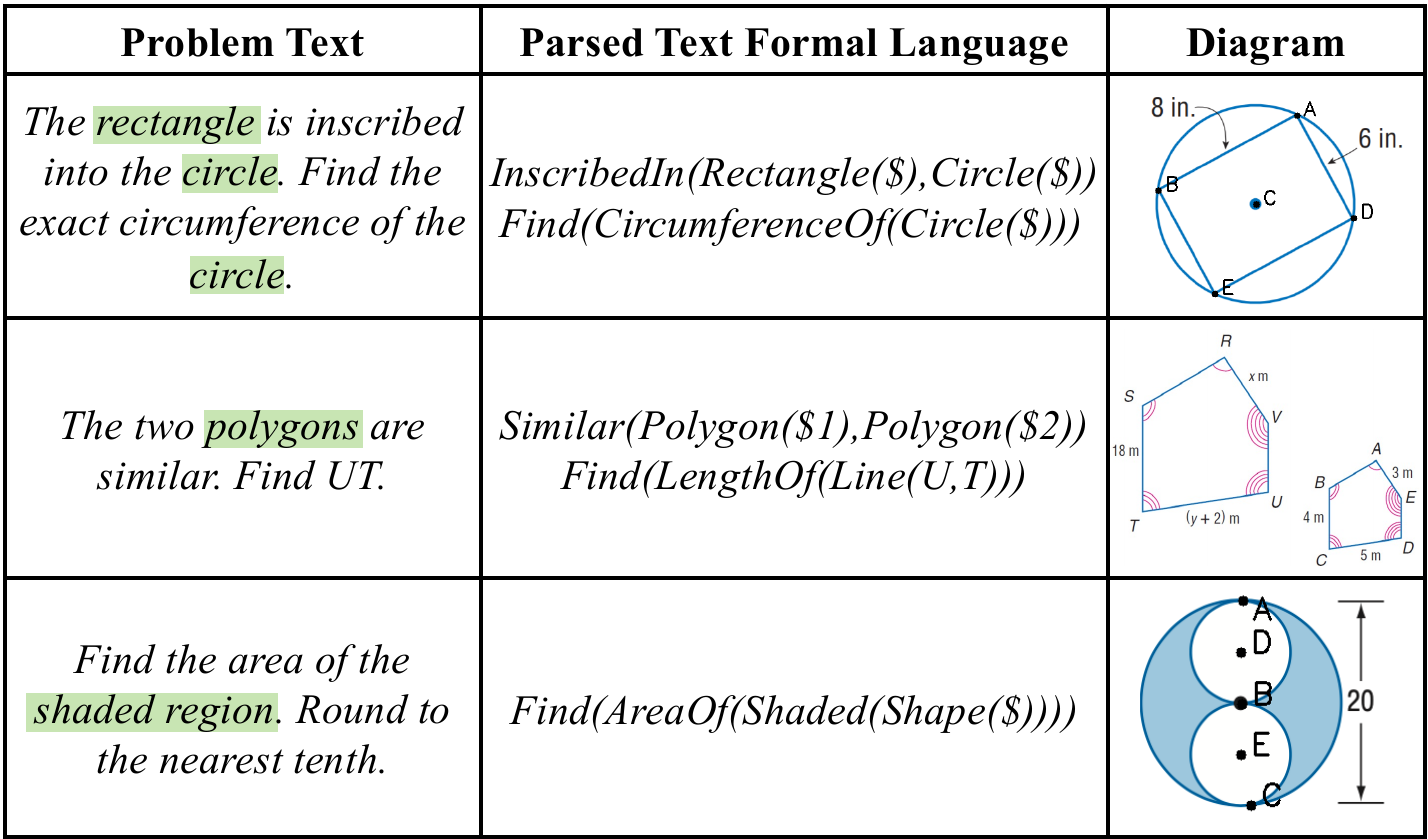}
  \caption{Illustrating the ambiguity presented in text. Text alone offers insufficient information to resolve the ambiguity, and disambiguation becomes straightforward when supported by a diagram.}
  \label{fig:intro}
  \vskip -0.2in
\end{figure}

Existing approaches to GPS can be broadly categorized into symbolic~~\cite{seo2015solving,alvin2017synthesis} and neural-based methods~~\cite{GeoQA,huang2020neural,Zhang2023PGPS,LANS,zhang2024fuse,xiao2024learning,Gsm-symbolic,GeoX,Eagle,gold,math-llava,Mavis,Math-puma,G-llava}. Symbolic methods, grounded in formal logic and mathematical rigor, rely on explicit theorem databases and symbolic manipulation to construct logically sound reasoning paths. These methods excel in providing interpretable steps and ensuring formal correctness. However, the predefined rules may struggle to accommodate diverse problem types.
In contrast, neural-based approaches leverage data-driven learning to generate solution paths from vast training datasets. These models offer flexibility and scalability, handling problems of varying complexity. However, their reliance on large, high-quality annotated datasets and the lack of rigorous correctness guarantees pose significant limitations.
Therefore, many works~~\cite{lu2021inter,Wu_2024_CVPR,peng-etal-2023-geodr,FGeo-DRL} attempt to combine the procedural power of symbolic models with the general power of neural models. Such hybrid approaches in general involves two key steps: parsing and reasoning. Parsing entails extracting formal language representations from the diagram and the accompanying text, while reasoning employs these parsed elements to predict and apply relevant theorem rules, ultimately constructing a logical path that leads to the final solution. This paper also builds on this emerging direction, offering novel insights about the pivotal role of diagrammatic information.

In this paper, we propose Pi-GPS, unleashing the power of diagrammatic information for enhancing geometry problem solving. Our work is inspired by the observation that text often conveys ambiguity in ways that diagrams, by nature, cannot easily accommodate~~\cite{1995cognitive}. However current approaches typically parse text and diagrams independently, resulting in ambiguities remain unresolved in text and further undermines the subsequent theorem prediction stage as the predictor's understanding of the problem is constrained. For instance, consider a text reference to "a shape." This could refer to a variety of geometric forms, such as a triangle, rectangle, or circle. Yet the text alone offers insufficient information to resolve the ambiguity. In contrast, we can easily disambiguate the reference when supported by a diagram, as the visual context clarifies the intended meaning.
Figure~\ref{fig:intro} provides several examples that highlight the ambiguities present in the text.

In light of this, our objective is to enhance geometry problem solving by introducing a micro module that resolves textual ambiguities through the diagrammatic information. We identify three primary sources of these ambiguities: (1) unspecified points (e.g., missing point names), (2) unspecified shapes (e.g., missing shape names), and (3) unspecified areas (e.g., computing shaded areas). To address these, we leverage Multimodal Large Language Models (MLLMs) to develop an error-correcting tool, rectifier, capable of automatically detecting and rectifying these ambiguities given the diagram as input.
Additionally, we design a verifier to mitigate MLLM's hallucination by verifying the disambiguated text aligns with diagrammatic heuristics (e.g., closed-loop shapes), which we show is pivotal in the experiments. We also explore the impact of recent advanced LLMs for reasoning, o3-mini~\cite{o3-mini}, in predicting theorem order, and present valuable analysis.

Experimentally we demonstrate our framework Pi-GPS, by resolving ambiguities in text, significantly outperforms state-of-the-art baselines on both Geometry3K~\cite{lu2021inter} and PGPS9K~\cite{Zhang2023PGPS} benchmarks.
We hope this work will draw attention to the crucial need for resolving text ambiguity in formal language space, an aspect often overlooked in previous research, and underscores its significance in advancing geometry problem solving.

In summary, our key contributions are:

% To address these challenges, we propose \textbf{Sync-GPS}, a novel framework for symbolic geometry problem-solving that combines advanced diagram parsing, rule-based text analysis, and a dedicated multimodal alignment module powered by MLLMs. Our framework aligns textual and visual elements in a fine-grained manner and employs rule-based mechanisms to validate alignment quality, thereby ensuring the explainability and mathematical rigor of the final solution process. This research not only provides a viable solution to the longstanding challenge of multimodal alignment in geometry problem-solving but also contributes a generalizable approach that can be applied to other domains requiring rigorous symbolic reasoning with visual-textual association. We believe Sync-GPS paves the way for further explorations of hybrid symbolic-neural methodologies in AI education and research applications. In particular:

\begin{itemize}[label=\textbullet]
    \item \textbf{Perspective.} We identify that text ambiguity is a key factor hindering the performance of geometry problem solving, which has been overlooked in prior works.

    \item \textbf{Methodology.} We propose a micro module to address text ambiguity, comprising a rectifier and a verifier. The rectifier powered by a MLLM refines text with diagrammatic information, while the verifier ensures alignment with diagrammatic heuristics. These components work in tandem to reduce ambiguity, which is further evaluated on theorem prediction using a strong reasoning LLM.
    
    \item \textbf{Evaluation.} The resulting framework, Pi-GPS, achieves the state-of-the-art performance, with a nearly 10\% improvement on Geometry3K over prior neural-symbolic approaches. We also provide strong evidence supporting the efficacy of the proposed module, and present an in-depth analysis.
\end{itemize}
\section{Related Works}
\begin{figure*}[t]
  \centering
  \includegraphics[width=\textwidth]{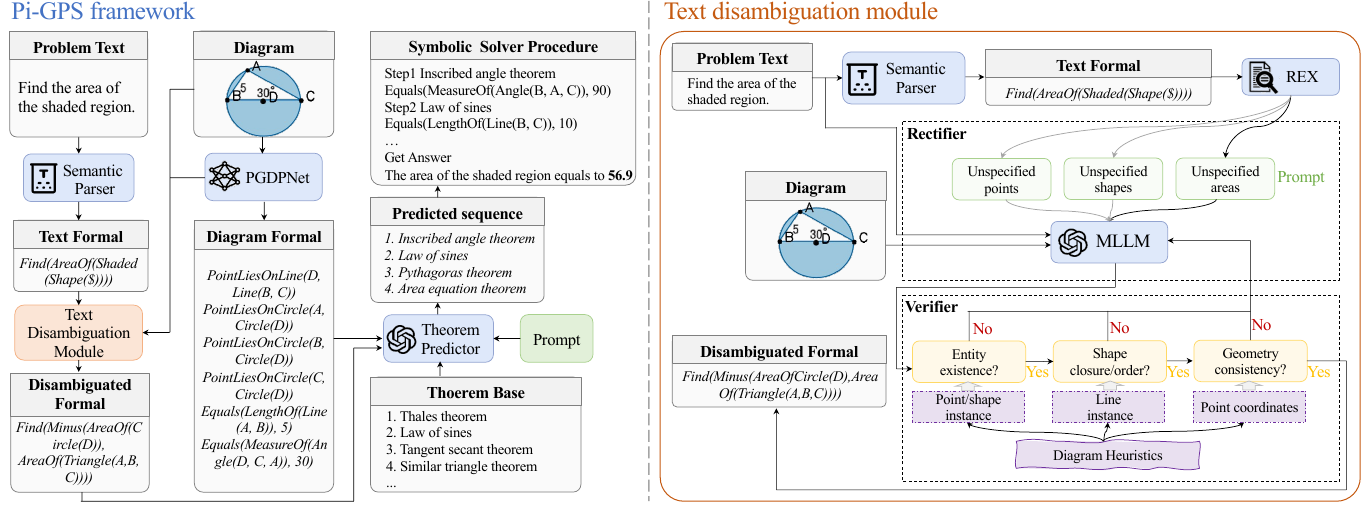}
  \caption{Illustrating the pipeline of our Pi-GPS: the overall framework is shown on the left and the text disambiguation module is depicted on the right, which plays a crucial role in resolving text ambiguity, enhancing performance. REX stands for regex pattern matching.}
  \label{fig:structure}
  \vskip -0.2in
\end{figure*}

\noindent
\textbf{Geometry Problem Solving.}
Recent advancements in automated GPS~\cite{sachan2017learning,seo2015solving,alvin2017synthesis} have attracted considerable attention due to the inherent complexity and unique challenges it presents. 
One prominent approach to GPS has been the use of language models that treat it as a specialized form of text generation. Notable examples include GeoQA~\cite{GeoQA} and UniGeo~\cite{UniGeo}. These models leverage large-scale pre-trained language models to generate solutions by interpreting geometry problems as text-based tasks. PGPS-Net~\cite{Zhang2023PGPS} improved upon these models by enhancing the performance of neural network-based approaches. These methods struggle to accurately capture the complex relationships between geometric entities in diagrams. 
LANS~\cite{LANS} circumvents this issue by incorporating diagram annotation, however, such annotations are not always accessible.  Additionally, the vector representations used by these models lack interpretability, resulting in unreliable or inconsistent solutions.
In contrast, symbolic systems approach GPS from a more structured, interpretable angle, such as 
GEOS~\cite{GeoS} and Inter-GPS~\cite{lu2021inter}.
They convert problem statements and diagrams into structured formats, enabling the application of symbolic solvers based on known geometric theorems. This approach enhances the interpretability of the problem solving process, yielding more precise solutions. 
A significant advancement in GPS was the introduction of PGDP~\cite{Zhang2022}, the first end-to-end diagram parsing method for geometry problems.
This was further used in works like GeoDRL~\cite{peng-etal-2023-geodr} and E-GPS~\cite{Wu_2024_CVPR}, which enhanced solution accuracy and robustness through techniques such as theorem library augmentation and theorem sequence prediction.
Our study introduces a novel micro module to resolve ambiguities in textual problem statements, which is orthogonal and can be plugged into existing neural-symbolic frameworks.

% for diagram-based text disambiguation. This module leverages contextual information from geometric diagrams to resolve ambiguities in textual problem statements.

% introduce a symbolic approach to enhance the interpretability of the problem-solving process by converting problem statements and schematic diagrams into formal language and employing symbolic solvers to apply theorems based on known relationships, thereby significantly improving solution precision and interpretability. Nonetheless, this method falls short in effectively managing the relationships between geometric elements within schematic diagrams. PGDP~\cite{Zhang2022} was the first to propose an end-to-end schematic diagram parsing method, markedly enhancing the capability to extract diagram formal language, which is further used in subsequent works GeoDRL~\cite{peng-etal-2023-geodr} and E-GPS~\cite{Wu_2024_CVPR}. The two methods pay more attention to theorem library augmentation and theorem sequence prediction. In our study, we also utilize PGDP as the diagram parser and introduce a new micro module to utilize the diagram for text disambiguiation. 

\noindent
\textbf{MLLMs for Mathematical Reasoning.}
% Research on MLLMs originates from investigating the effective integration of natural language processing and computer vision, enabling them to excel in tasks requiring both visual and linguistic comprehension. 
Early research in multimodal learning focused on leveraging attention mechanisms to align image and text representations. A key breakthrough came with CLIP~\cite{clip}, which learned transferable visual representations through natural language supervision, laying the foundation for subsequent large-scale multimodal models. Building on this, the LLaVA series~\cite{LLaVA, math-llava, G-llava} introduced visual instruction-tuning, linking a visual encoder to a language model via a simple multi-layer perceptron.
Subsequent works~\cite{Flamingo,Qwen-VL,Deepseek-vl,Qwen2-vl} expanded MLLMs by incorporating novel visual perception modules and hybrid vision encoders, enhancing their ability to address increasingly complex tasks. As model scale has grown, so too has their ability to perform sophisticated contextual and mathematical reasoning~\cite{GPT-f, gsm8k, mathqa}. More recently, models like MathGLM-Vision~\cite{MathGLM-Vision} and Math-LLaVA~\cite{math-llava} have introduced chain-of-thought reasoning and intermediate-step generation, enabling problem solving by breaking down complex tasks into manageable steps.
Despite their impressive contextual reasoning abilities, MLLMs still face challenges such as hallucination, which is particularly problematic in mathematical reasoning tasks. In our work, we utilize MLLMs to generate diagrammatic information for text disambiguation, framing this as a more tractable task within the zero-shot capabilities of these models.

% For instance, GPT-f~\cite{GPT-f} demonstrates outstanding performance in both mathematical and logical reasoning tasks. 
% Meanwhile, research on multimodal mathematical reasoning has evolved from relying solely on text (as exemplified by datasets such as GSM8K and MathQA) to more intricate and challenging tasks that demand handling textual, mathematical, and visual information simultaneously. This shift underscores the significance of integrating natural language explanations with visual cues in tackling problems involving geometry, graphical sketches, and proof derivations. 

\section{Method}

In geometry problem solving, the \emph{Diagram} represents the geometric figure, while the \emph{Problem Text} provides the textual description, including the problem's objective (e.g., "Find the length of EH"). The goal is to solve for the correct answer corresponding to the given pair. To achieve this, we propose \textbf{Pi-GPS}, as illustrated in Figure~\ref{fig:structure}.
The framework comprises a parser and a reasoner. Distinct from previous symbolic approaches, we propose a text disambiguation module that leverages the diagrammatic information to resolve text ambiguity, and we introduce a theorem predictor using an LLM given the disambiguated formal language.
% Our method first converts the textual problem description into a formal language (Section ~\ref{subsec:Text Parser}), and then transfers the diagram into a formal language (Section ~\ref{subsec:Diagram Parser}).

\subsection{Parser}

\noindent 
\textbf{Text Parser.}
A critical step in solving geometry problems is extracting relevant information from the problem statement, particularly identifying the premises and the goal of the problem. This extraction process can be categorized into rule-based methods and deep neural network-based methods. Traditional rule-based parsing techniques have been shown to provide relatively precise results. Although deep neural networks excel in sequence-to-sequence (Seq2Seq) tasks such as machine translation, previous research~\citep{lu2021inter} suggests that Seq2Seq-based semantic parsers struggle with geometry problems. This is primarily due to the limited size of geometric datasets and the tendency of neural parsers to introduce noise into the output.

To achieve more accurate parsing, we utilize a rule-based text parser following~\citep{lu2021inter,Wu_2024_CVPR,peng-etal-2023-geodr}. This parser analyzes the problem text by applying regular expressions to identify basic elements, numerical values, and their interrelationships. The parser automatically generates a set of propositions \(P_T\) and identifies the problem target \(t^*\) from the text.

\noindent
\textbf{Diagram Parser.}
As with previous work~\cite{Wu_2024_CVPR,peng-etal-2023-geodr}, we employ PGDPNet~\citep{Zhang2022}, an end-to-end neural network-based model that efficiently extracts basic elements such as points, lines, and circles from geometric figures along with their logical relationships, and generates a formal set of propositions. This approach achieves state-of-the-art performance in geometric diagram parsing.

\begin{figure}[t]
  \centering
  \includegraphics[width=\columnwidth]{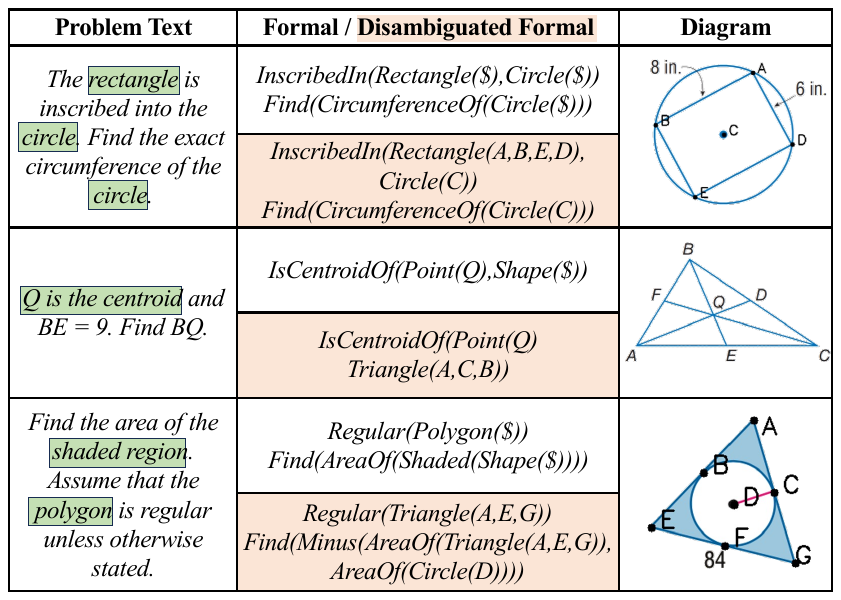}
  \caption{Illustrating several examples, showing the proposed text disambiguation module is capable of resolving text ambiguity.}
  \label{fig:aligned}
  \vskip -0.2in
\end{figure}

\noindent
\textbf{Text Disambiguation Module.}
Previous studies~\citep{peng-etal-2023-geodr, Wu_2024_CVPR} have typically concatenated the parsing results from text and diagram parsers in a straightforward manner. While each parser performs well within its respective modality, the disconnect between textual and visual representations often leads to unresolved ambiguities, impeding subsequent solving process and thus degrading the overall accuracy. To overcome this issue, we propose a novel module incorporating a rectifier and a verifier, as illustrated in Figure~\ref{fig:structure}.

\noindent
\textbf{(i) Rectifier using MLLM.}
% In previous related works, alignment modules can be categorized into neural network-based methods and those leveraging multi-modal large models. Neural network approaches require substantial amounts of labeled training data, and the outputs from modal fusion typically necessitate additional transformation to achieve the desired formalized language. Conversely, multi-modal large models demonstrate exceptional performance in aligning visual and textual information. They are adept at generating formalized text and exhibit strong generalization capabilities, eliminating the need for extra training and allowing effective completion of various tasks through appropriate prompting.
The primary objective of the rectifier is to resolve ambiguities in the output of the text parser by leveraging the diagram through a MLLM. 
Upon analyzing the sources of ambiguity, we categorize the root causes into three distinct types, which are as follows:
\begin{itemize}
    \item \textbf{Unspecified points}:  
    The text parser can identify specific geometric shape but fail to associate them with explicit points. For example, in the relationship \texttt{CircumscribedTo(Square(\$), Circle(\$))}, the parser recognizes that a square is circumscribed to a circle but does not specify the defining points (vertices) of the square or circle.
    % we provide the MLLM with the extracted shape categories from the text parser. 
    % We prompt the MLLM to output the corresponding letters for the queried shapes, ensuring accurate alignment between textual descriptions and diagrammatic representations.
    
    \item \textbf{Unspecified shapes}:  
    The text parser, although capable of recognizing certain geometric constructs, fails to correctly map or associate them with predefined shapes or geometric entities in the system. This limitation is evident in expressions like \texttt{IsAltitudeOf(Line(C,P), Shape(\$))}, where the absence of explicit shape identification impedes effective interpretation.
    
    % we supply the MLLM with additional prompts containing formal language expressions for various shapes. This requires the MLLM to simultaneously output both the graphical shape and the corresponding letter as referenced in the formal language, thereby bridging the gap between textual and visual data.
    
    \item \textbf{Unspecified areas}:  
    The text parser indicates that the formal language specifies graphical elements, such as \texttt{Find(AreaOf(Shaded(Shape(\$)))}. This implies the need to determine the area of a shaded region, typically representing areas of interest within a diagram.
    % Yet, the lack of explicit boundaries or additional descriptors within the language necessitates a robust approach to infer the precise location, extent, and properties of these areas.

% A comprehensive solution must account for various possibilities, including different types of shading (e.g., solid, gradient, patterned) and the role of these visual elements in the larger context of the problem. To address this, advanced algorithms may be employed to dynamically interpret the text-parser output, mapping it to corresponding geometric constructs within a coordinate system. Furthermore, methods for resolving ambiguities—such as determining the boundaries of shaded regions or identifying intersecting areas—are critical in ensuring the accurate calculation of such regions' areas.

% By developing techniques that can automatically formalize and solve these issues, we can enhance the parser’s ability to handle complex, visually represented problems and translate them into solvable geometric tasks.
    
    % we prompt the MLLM with specialized notation methods for shapes (e.g. \texttt{Minus(Shape(A),Shape(B))}, \texttt{Add(Shape(A),Shape(B))}). This guides the MLLM to correctly represent the corresponding special shapes, ensuring precise alignment and accurate formalization of complex graphical information.
\end{itemize}

The rectification process begins by employing regular expressions to identify unknown identifiers, represented by the symbol '\$', and determine the type of ambiguity. For each identified ambiguity, a specific prompt is crafted based on its nature, and the MLLM is used to resolve the issue by referencing both the \emph{Diagram} and \emph{Problem Text}. The incorporation of diagrammatic information is crucial, as it provides supplementary context to improve resolution accuracy. 
% This approach ensures that a broad range of ambiguity can be addressed, leveraging the strengths of both textual and diagrammatic inputs. 
However, the potential for hallucination must be carefully managed, as it could compromise the accuracy of the rectification. 
Meanwhile, generating output in a formal language poses a significant challenge for MLLMs, as absolute precision is essential, any deviation such as an incorrect character and misplaced parenthesis can invalidate the output.
This motivates us to design a verifier based on diagram heuristics, ensuring its correctness.

\noindent 
\textbf{(ii) Verifier using Diagram Heuristics.}
We propose a Logical Reasoning Verifier that utilizes diagram heuristics derived from the diagram parser to ensure the consistency of outputs generated by the MLLM with the provided geometric diagram. This verifier incorporates three key heuristics for examining the rectified output:
\begin{itemize}
\item \textbf{Entity existence verification.} MLLMs may generate geometric entities such as points, lines, or circles that do not correspond to actual elements in the diagram. We use the entity instances identified by the diagram parser to cross-check the MLLM's output, ensuring consistency with the original diagram.
\item \textbf{Shape closure and order validation.} A common issue occurs when points within a geometric shape fail to form a closed figure or are ordered incorrectly. For instance, a pentagon labeled Pentagon(A,B,D,E,C) may be erroneously output as Pentagon(A,B,C,D,E). To rectify this, we construct a graph representing the diagram, checking for connectivity and cyclic properties, and ensure each node has the correct degree for valid closure. If the points are ordered incorrectly, we reorder the vertices based on the graph structure to ensure proper shape formation.
\item \textbf{Geometry consistency of vertices.} The MLLM-generated vertices may not always align with the intended geometry shape. We apply analytical geometry techniques to verify that the vertices match the intended shape based on their coordinates.
\end{itemize}

When discrepancies arise, feedback from the verifier is incorporated into the rectifier, creating a loop that allows the MLLM to iteratively adjust and refine its output.
Experimental results demonstrate that the verifier plays a crucial role in ensuring adherence, thereby enhancing geometric problem solving accuracy.

\subsection{Reasoner}\label{subsec:predictor}
The reasoner typically comprises a predictor for theorem order prediction and a solver that applies the theorems in the predicted order to derive the final solution.

\begin{table*}[t]
\centering
\footnotesize % 调整字体大小
\begin{tabularx}{\textwidth}{   
    p{0.14\textwidth} 
    >{\centering\arraybackslash}p{0.06\textwidth} 
    >{\centering\arraybackslash}X
    *{9}{>{\centering\arraybackslash}p{0.05\textwidth}} 
}
\toprule
\multirow{2}{*}{\textbf{Methods}} & 
\multirow{2}{*}{\textbf{Accuracy}} & 
\multirow{2}{*}{\textbf{Steps}} & 
\multicolumn{4}{c}{\textbf{Question Type}} & 
\multicolumn{5}{c}{\textbf{Geometric Shape}} \\
\cmidrule(lr){4-7} \cmidrule(lr){8-12}
& & & Angle & Length & Area & Ratio & Line & Triangle & Quad & Circle & Other \\
\midrule
Human~\citep{lu2021inter} & 56.9 & – & 53.7 & 59.3 & 57.7 & 42.9 & 46.7 & 53.8 & 68.7 & 61.7 & 58.3 \\
Human Expert~\citep{lu2021inter} & 90.9 & – & 89.9 & 92.0 & 93.9 & 66.7 & 95.9 & 92.2 & 90.5 & 89.9 & 92.3 \\
\hline
Gemini 2~\citep{gemini} & 60.7 & – & 58.9 & 61.8 & 57.5 & 68.8 & 54.1 & 62.7 & 45.5 & 57.7 & 58.3 \\
Claude3.5~Sonnet~\citep{claude35} & 56.4 & – & 54.9 & 57.3 & 53.6 & 64.6 & 49.4 & 58.6 & 40.9 & 57.9 & 53.9 \\
GPT-4o~\citep{gpt4}&58.6 & – & 55.6 & 59.3 & 55.1 & 70.6 & 51.4 & 60.4 & 43.1 & 59.0 & 56.7 \\
\hline
Inter-GPS~\citep{lu2021inter} & 57.5 & 7.1 & 59.1 & 61.7 & 30.2 & 50.0 & 59.3 & 66.0 & 52.4 & 45.5 & 48.1 \\
GeoDRL~\citep{peng-etal-2023-geodr} & 68.4 & – & 75.5 & 70.5 & 22.6 & 83.3 & 77.8 & 76.0 & 62.9 & 59.4 & 48.1 \\
E-GPS~\citep{Wu_2024_CVPR} & 67.9 & 1.63–2.28 & 78.3 & 67.2 & 27.7 & 72.2 & 76.1 & 75.6 & 59.4 & 55.0 & 51.8 \\
Pi-GPS (ours) & \textbf{77.8} & 2.31–4.12 & \textbf{83.9} & \textbf{81.4} & \textbf{59.0} & \textbf{81.2} & \textbf{79.6} & \textbf{83.9} & \textbf{76.4} & \textbf{73.0} & \textbf{69.4} \\
\bottomrule
\end{tabularx}
\caption{Comparison of geometry problem solving on the Geometry3K dataset. Our method consistently outperforms all baseline models. Accuracy, Steps, and additional metrics are reported for different question types and geometric shapes. Best results are highlighted in bold.}
\label{tab:geometry3k_performance}
\vskip -0.2in
\end{table*}

\noindent
\textbf{Theorem predictor.}
% The prediction of theorems is a crucial step in Geometry Problem Solving process. 
Accurately predicting the correct sequence of theorems is essential for deriving solutions and ensuring the interpretability. Previous approaches~\citep{lu2021inter, Wu_2024_CVPR} have used transformer-based models, framing theorem sequencing as a sequential prediction task. Additionally, some study~\citep{peng-etal-2023-geodr} has applied reinforcement learning to improve theorem prediction accuracy.
However, a major limitation of these methods is their reliance on annotated problem solving sequences for training, which are often scarce or expensive to generate, particularly in specialized domains requiring domain-specific expertise.
Building on recent advancements, we draw inspiration from AlphaGeo~\citep{AlphaGeo}, which demonstrates the potential of LLMs in symbolic deduction. In this work, we explore the application of advanced LLMs, specifically the o3-mini~\cite{o3-mini}, by prompting the model with a library of geometry theorem knowledge. The model then generates the most appropriate order of theorems based on disambiguated text and diagram formals. This approach reduces dependence on labeled data and leverages the generalization capabilities of modern LLMs.

% By incorporating this methodology, we aim to enhance both the accuracy and interpretability of theorem prediction processes, providing a more efficient framework for solving geometric problems that require rigorous step-by-step logical reasoning.

% This strategy not only ensures the accuracy of the results but also achieves strong generalization. To further evaluate the effectiveness of the alignment module, we incorporate a traversal strategy to maximize the correctness of theorem applications. However, to ensure the interpretability of the final sequence of theorems, we verify the validity of each theorem during traversal and apply it only after confirming that it contributes effectively to problem solving. 

% Consequently, verified randomly generated sequences are frequently employed as pseudo-optimal sequences, which may introduce uncertainty in prediction performance. however, this approach demands extensive training and is highly dependent on the size of the theorem library, thereby limiting its generalizability.
\noindent
\textbf{Solver.}
Our solver framework builds upon the approach in~\citep{lu2021inter} and incorporates the expanded theorem library from~\citep{peng-etal-2023-geodr}. We have modified the logical framework to accommodate the extended formal language, specifically to address shadow regions and other special cases. The solver, along with the extended theorem library, is also employed in the experimental baselines for a fair comparison.

\section{Experiments}

\begin{table*}[t]
\centering
\small % 减小字体
\begin{tabularx}{\textwidth}{p{0.18\textwidth} X 
>{\centering\arraybackslash}p{0.11\textwidth} 
>{\centering\arraybackslash}p{0.11\textwidth} 
>{\centering\arraybackslash}p{0.11\textwidth} 
>{\centering\arraybackslash}p{0.11\textwidth}}
\toprule
\multirow{2}{*}{Category} & \multirow{2}{*}{Method} & \multicolumn{2}{c}{Geometry3K} & \multicolumn{2}{c}{PGPS9K} \\
\cmidrule(lr){3-4} \cmidrule(lr){5-6}
 &  & Completion & Choice & Completion & Choice \\
\midrule
\multirow{4}{*}{MLLMs} 
 & Qwen-VL~\citep{Qwen-VL} & 22.1 & 26.7 & 20.1 & 23.2 \\
 & GPT-4o~\citep{gpt4} & 34.8 & 58.6 & 33.3 & 51.0 \\
 & Claude 3.5 Sonnet~\citep{claude35} & 32.0 & 56.4 & 27.6 & 45.9 \\
 & Gemini 2~\citep{gemini} & 38.9 & 60.7 & 38.2 & 56.8 \\
\midrule
\multirow{6}{*}{Neural Methods} 
 & NGS~\citep{GeoQA} & 35.3 & 58.8 & 34.1 & 46.1 \\
 & Geoformer~\citep{geoformer} & 36.8 & 59.3 & 35.6 & 47.3 \\
 & SCA-GPS~\citep{SCA-GPS} & - & 76.7 & - & - \\
 & GOLD$^*$~\citep{gold} & - & 62.7 & - & 60.6\\
 & PGPSNet-v2-S$^*$~\citep{zhang2024fuse} & 65.2 & 76.4 & 60.3 & 69.2 \\
 & LANS (Diagram GT)$^*$~\citep{LANS} & 72.1 & 82.3 & 66.7 & 74.0 \\
\midrule
\multirow{4}{*}{Neural-symbolic Methods} 
 & Inter-GPS~\citep{lu2021inter} & 43.4 & 57.5 & - & - \\
 & GeoDRL~\citep{peng-etal-2023-geodr} & 57.9 & 68.4 & 55.6 & 66.7 \\
 & E-GPS~\citep{Wu_2024_CVPR} & - & 67.9 & - & - \\
 & Pi-GPS (ours) & \textbf{70.6} & \textbf{77.8} &  \textbf{61.4} & \textbf{69.8} \\
\bottomrule
\end{tabularx}
\caption{\label{tab:methods_performance}
Comparison of geometry problem solving on Geometry3K and PGPS9K. Our method achieves the best performance (highlighted in bold) compared to the neural-symbolic methods. Note that LANS relies on textual clauses and point positions derived from diagram annotations. '*' is to denote that the decoders of PGPSNet and LANS are trained on the larger dataset, PGPS9K.
}
\vskip -0.1in
\end{table*}

\subsection{Settings}
\noindent
\textbf{Datasets.}
We conduct experiments using the Geometry3K~\citep{lu2021inter} and PGPS9K~\citep{Zhang2023PGPS} datasets. Geometry3K consists of 3,002 geometry problems, partitioned into 2,101 for training, 300 for validation, and 601 for testing. Each problem is accompanied by a geometric diagram, problem text, and formal language parsing annotations. It covers a diverse range of geometric shapes, including lines, triangles, circles, quadrilaterals, and other polygons, making it a comprehensive benchmark. PGPS9K, an expanded version of Geometry3K, contains 9,022 geometry problems paired with 4,000 unique diagrams. Of these, 2,891 problems with 1,738 diagrams are sourced from Geometry3K, while the remaining problems are collected from five widely-used mathematics textbooks for grades 6-12, covering nearly all plane geometry problem types for these educational levels.

\noindent
\textbf{Metrics.}
Building on the methodologies of prior studies~\citep{lu2021inter,Zhang2023PGPS}, we adopt two evaluation schemes: \textit{Completion} and \textit{Choice}, to assess the numerical performance of our methods. The \textit{Completion} metric gauges the model’s ability to generate the first executable solution program as its final output. The \textit{Choice} metric measures the model's ability to correctly select an option from four candidates, with random selection as a fallback when the generated answer does not match any provided options. Performance is evaluated based on accuracy.

% Additionally, we assess the performance of the latest MLLMs. In Completion mode, the MLLM is required to generate answers directly without additional prompts. In Choice mode, reference options are included in the prompt to guide the large model in selecting the appropriate answer.

\noindent
\textbf{Baselines.}
We conduct a comprehensive comparison between our proposed method and state-of-the-art models across various categories to analyze their performance in geometry problem solving tasks. For neural solvers, we evaluate several prominent models: NGS~\citep{GeoQA}, which uses a ResNet-101 architecture for encoding geometric diagrams; Geoformer~\citep{geoformer}, which employs the VL-T5 model for diagram encoding followed by a Transformer-based processing architecture; SCA-GPS~\citep{SCA-GPS}, which introduces a novel strategy for geometric problem-solving; PGPSNet~\citep{Zhang2023PGPS}, which combines CNN and GRU encoders to enhance geometric reasoning;
LANS~\cite{LANS}, a layout-aware neural solver;
as well as GOLD~\cite{gold}, which converts geometry diagrams into natural language descriptions.
In the realm of neural-symbolic solvers, we compare with the classical Inter-GPS~\cite{lu2021inter} and two advanced models: GeoDRL~\cite{peng-etal-2023-geodr}, which improves Inter-GPS's search strategy by integrating logical graph deduction and deep reinforcement learning; and E-GPS~\cite{Wu_2024_CVPR}, which combines top-down and bottom-up reasoning to match the performance of other methods with fewer steps and improved explainability.
Additionally, we report results from leading MLLMs, including Qwen-VL~\cite{Qwen-VL}, GPT-4o~\citep{gpt4}, Gemini 2~\citep{gemini}, and Claude 3.5 Sonnet~\citep{claude35}, which represent cutting-edge visual reasoning capabilities.
It is important to note that our method does not require ground-truth parsing (neither diagram annotation nor text annotation). 
% Consequently, all models in this comparison are evaluated without external annotations, ensuring a fair and consistent benchmark across methods.
We adopt the expanded theorem set from GeoDRL, which is also utilized by other methods that require a theorem base for fair comparison.

\subsection{Results}
We present a detailed comparison of our method with both MLLMs and neural-symbolic baselines on the Geometry3K dataset, as summarized in Table~\ref{tab:geometry3k_performance}. Our method consistently outperforms all baseline models, demonstrating superior performance, and even surpassing human experts in certain subcategories, such as the ratio question type.
While MLLMs excel in general multimodal tasks, they exhibit limitations when applied to specialized mathematical geometry problems. These challenges stem from MLLMs' difficulty in accurately parsing geometric diagrams, performing complex reasoning, and executing precise numerical computations. In contrast, our method not only achieves significantly better results but also offers greater interpretability.
Compared to neural-symbolic baselines, our approach achieves the highest performance, with an impressive improvement of nearly 10\% over the two strong baselines, E-GPS and GeoDRL. This improvement highlights the substantial impact of text ambiguity, an often overlooked factor in prior work. 
% Ambiguity in the problem text impedes a model's comprehension, thus limiting its accuracy. 
Our category analysis reveals that text ambiguity affects all categories, with the most significant impact observed in the area question type, where the text frequently refers to "the area" without a specific identifier, exacerbating the ambiguity.

To further validate the effectiveness of our approach, we present additional comparisons on the PGPS9K dataset for both completion and choice evaluation tasks, as shown in Table~\ref{tab:methods_performance}. Notably, compared to the state-of-the-art neural-symbolic method, GeoDRL, our method achieves improvements of 5.8\% and 3.1\% on the PGPS9K dataset in terms of completion and choice respectively. These results highlight the efficacy of our approach in interpreting the semantic intent of problem statements, a capability enabled by the integration of our text disambiguation module.
Additionally, we provide a comprehensive comparison with competitive neural methods. Notably, LANS~\cite{LANS}, the top-performing neural model, relies on textual clauses and point positions derived from diagram annotations. This reliance on ground-truth annotations significantly boosts performance. Both LANS and PGPSNET were trained on the large-scale PGPS9K dataset. However, neural-based methods typically suffer from a lack of interpretability. In contrast, our method advances the field of interpretable geometry problem solving by addressing the critical issue of text ambiguity, a challenge often overlooked in previous work.

\subsection{Analysis}

We provide an in-depth analysis about the framework of our Pi-GPS by conducting comprehensive ablation studies on Geometry3K dataset.

\begin{table}[t]
\centering
\small
\setlength{\tabcolsep}{4pt}
\begin{tabular}{ccccc}
\toprule
 Text disam. & Theorem pred. & Completion & Choice & Steps \\
\midrule
&  & 60.7 & 70.6 & 2.85-6.03\\
\checkmark & & 68.9 & 76.6 & 2.85-6.03 \\
& \checkmark & 63.2 & 72.3 & 2.31-4.12\\
\checkmark & \checkmark & \textbf{70.6} & \textbf{77.8} & \textbf{2.31-4.12}\\
\bottomrule
\end{tabular}
\caption{Illustrating the effect of text diambiguation module (Text disam.) and theorem predictor (Theorem pred.) on Geometry3K. The text disambiguation module plays a critical role with its  especially significant impact in driving performance improvement.}
\label{tab:Pi-GPS}
\vskip -0.2in
\end{table}

\noindent
\textbf{Text disambiguation module is pivotal in Pi-GPS.}
We first conduct an ablation study to analyze the impact of two key components in Pi-GPS: the text disambiguation module and the theorem predictor utilizing an LLM. When the theorem predictor is disabled, a traversal strategy is employed. The comparison results are presented in Table~\ref{tab:Pi-GPS}.
Notably, the superior baseline performance relative to prior work is largely attributed to our self-trained PGDP model, which may exhibit enhanced diagram parsing capability. However, even with this strong baseline, both proposed components continue to significantly improve performance.
Specifically, by integrating the proposed theorem predictor, the number of solving steps is reduced. More importantly, the text disambiguation module plays a critical role in Pi-GPS, with its impact being especially pronounced in driving overall performance improvement, a consistent enhancement of over 5\% across all cases.
% This highlights the pivotal role of textual disambiguation in the effectiveness of Pi-GPS.

% In our work, we establish the baseline model by adopting the primary framework of Inter-GPS along with the theorem set from GeoDRL. The enhancements observed, relative to previous methods, can be primarily attributed to the rectification of code errors in the original framework and the superior diagram parsing performance of our self-trained PGDP model.

\noindent
\textbf{The verifier is critical in text disambiguation module.}
We further examine the roles of the rectifier and verifier components within the text disambiguation module. The experimental results, summarized in Table~\ref{tab:text-dis}, also report the influence of a tailored prompt designed to address specific text ambiguity scenarios identified via regular expression (regex) pattern matching.
A key observation is that, without a tailored prompt, applying the general rectifier degrades performance due to hallucinations and uncertainties inherent in MLLMs. These factors introduce erroneous modifications, reducing disambiguation accuracy. In contrast, incorporating a tailored prompt improves performance beyond the baseline, underscoring the importance of domain-specific guidance in enhancing the rectifier’s effectiveness. This suggests that explicit contextual cues help mitigate unintended alterations and improve rectification precision.
Additionally, the verifier significantly enhances disambiguation, yielding performance gains of 4\%–6\%. This highlights its critical role in enforcing consistency and correctness by systematically validating and refining rectified outputs. These findings collectively demonstrate that effective coordination between rectification and verification, along with domain-specific prompting, is essential for robust and accurate text disambiguation.

\begin{table}[t]
\centering
\begin{tabular}{lcc}
\toprule
 Method & Completion & Choice \\
\midrule
Ours w/o Text disam.      & 63.2 & 72.3 \\
~~~+ Rectifier (general prompt)        & 62.4 & 71.9 \\
~~~+ Rectifier (specific prompt)  & 64.2 & 73.3 \\
~~~~~~+ Verifier  & \textbf{70.6} & \textbf{77.8} \\
\bottomrule
\end{tabular}
\caption{Illustrating the roles of the rectifier and verifier in the text disambiguation module on Geometry3K.}
\label{tab:text-dis}
\vskip -0.2in
\end{table}

\begin{table}[t]
\centering
\small
\setlength{\tabcolsep}{0.5pt}
\begin{tabular}{c|ccc}
\hline
Task & Models & Completion & Choice \\
\hline
\multirow{2}{*}{Direct solv. (MLLM)} & GPT-4o & 34.8 & 58.6\\
& Gemini 2 & 38.9 & 60.7\\
\hline
\multirow{4}{*}{Direct solv. (LLM)} & GPT-4o & 36.5& 59.7\\
% & Gemini2 & & \\
& DeepSeek-R1 & 63.9& 72.2\\
& o3-mini& 66.4 & 75.5 \\
\cline{2-4}
& o3-mini w/o Text disam. & 61.4 & 70.4 \\
\hline
Theorem pred. (LLM) & o3-mini (ours) & \textbf{70.6} & \textbf{77.8} \\
\hline
\end{tabular}
\caption{Illustrating different roles of (M)LLMs in GPS. While advanced LLMs exhibit strong mathematical reasoning in direct solution generation, our approach, leveraging LLMs for theorem prediction, improves both performance and interpretability.}
\label{tab:o3-mini}
\vskip -0.2in
\end{table}

\noindent
\textbf{Theorem order prediction is a more manageable task for LLMs.}
In our framework, we integrate an LLM to facilitate theorem prediction. A natural baseline is to directly employ an LLM to solve problems using the parsed formal representations. Given the growing interest in LLMs for mathematical reasoning, this approach warrants thorough investigation.
We conduct experiments and evaluate LLMs and MLLMs on their ability to directly solve the problem. The results are summarized in Table~\ref{tab:o3-mini}. 
We have several key observations.
(1)
\textbf{MLLMs struggle with geometry reasoning from raw inputs.} When treated as LLMs and provided with parsed text and diagram formal representations, MLLM, GPT-4o in this case, outperform their direct processing of original problem text and diagrams. This suggests that current MLLMs face challenges in extracting logical information from visual diagrams.
(2)
\textbf{o3-mini exhibits superior reasoning capabilities.} Among the evaluated LLMs, o3-mini consistently achieves the best performance when directly applied to problem solving, reaffirming its effectiveness in mathematical reasoning tasks.
(3)
\textbf{Ambiguities in parsed input significantly degrade performance.} When tested with ambiguous parsed text and diagram representations, o3-mini's accuracy drops substantially. This again validates our observation that disambiguating textual input is important to enhance the model’s reasoning capabilities, as even strong LLMs struggle with unresolved ambiguities in mathematical relationships.
(4)
\textbf{Theorem prediction enhances both accuracy and interpretability.} Rather than directly solving the problem, our method leverage LLMs to predict a sequence of theorems from a predefined theorem base, followed by a dedicated solver. This structured approach not only improves accuracy but also enhances interpretability, which is particularly valuable in educational settings where step-by-step justifications are crucial.

\noindent
\textbf{The effect of different LLMs in theorem predictor.}
We further conduct an ablation study on different LLMs used in our theorem predictor, with results on the Geometry3K dataset presented in Table~\ref{tab:theorem_results}. Compared to the vanilla traversal-based approach, incorporating LLMs improves solving accuracy while reducing the number of steps required, thereby enhancing overall efficiency. Notably, all evaluated LLMs achieve comparable performance, suggesting that theorem order prediction is a well-developed application for LLMs, demonstrating their robustness and reliability in this task.

\begin{figure}[t]
  \centering
  \includegraphics[width=1.0\columnwidth]{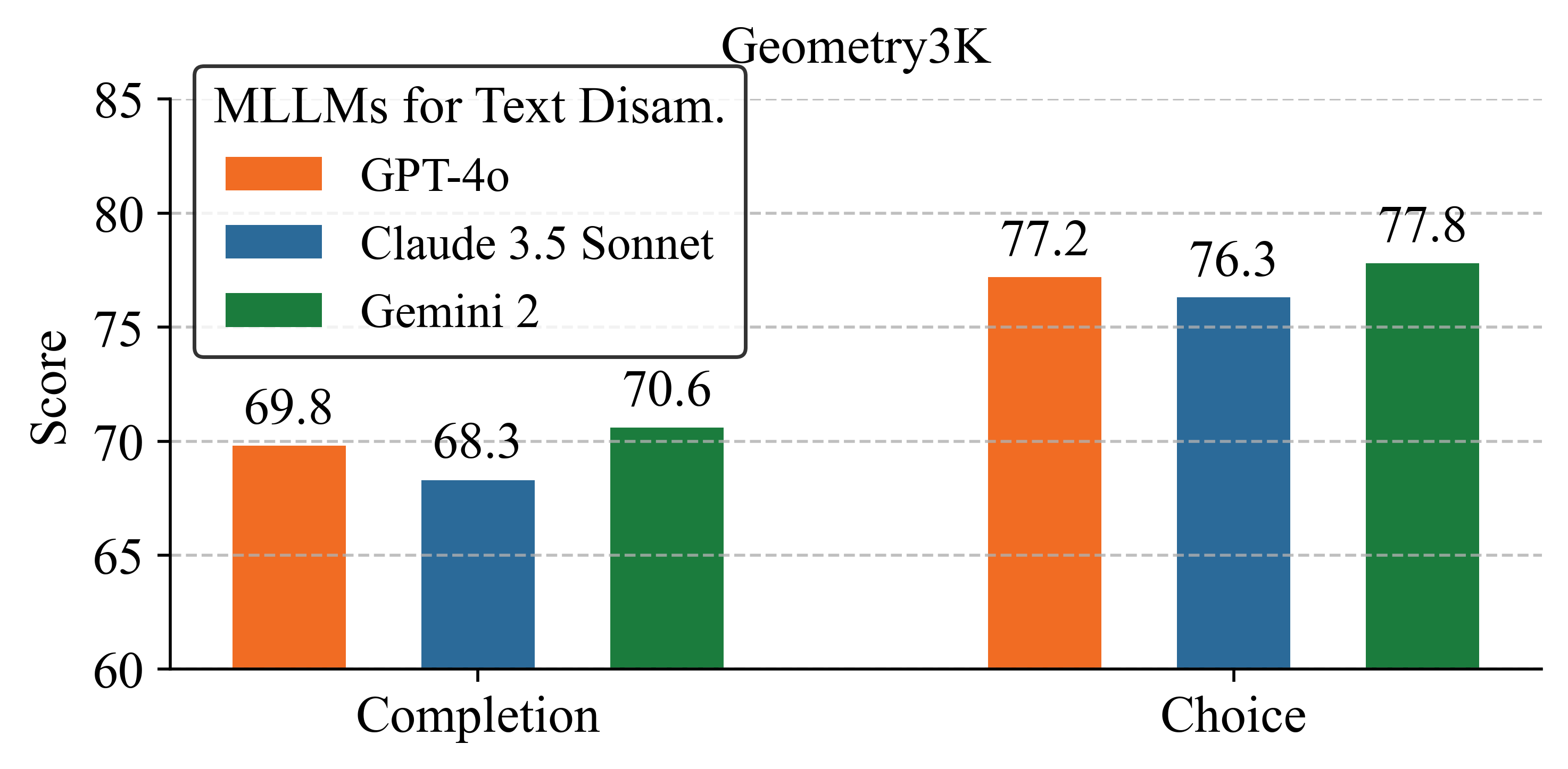}
  \caption{Illustrating the effect of different MLLMs used in rectifier within text disambiguation module.}
  \label{fig:MLLMs}
\end{figure}

\noindent
\textbf{The effect of different MLLMs in rectifier.}
In our approach, the rectifier within the text disambiguation module utilizes MLLMs to enhance performance. To explore the impact of different MLLMs on the rectification process, we conduct an ablation study evaluating multiple well-established MLLMs and present the results in Figure~\ref{fig:MLLMs}. The findings indicate that all tested models,  regardless of their inherent capabilities and design variations, yield substantial improvements over the baseline. This demonstrates that our method's effectiveness is not dependent on a specific MLLM but rather highlights its robustness and broad applicability across diverse architectures, reinforcing its generalizability.

% \begin{table}
% \centering
% \begin{tabular}{lcc}
% \toprule
% \multirow{2}{*}{Method} & \multicolumn{2}{c}{Geometry3K} \\
% \cmidrule(lr){2-3}
%  & Completion & Choice \\
% \midrule
% without align     & 63.2 & 72.3 \\
% GPT-4o            & 69.8 & 77.2 \\
% Claude 3.5 Sonnet & 68.3 & 76.3 \\
% Gemini 2          & 70.6 & 77.8 \\
% \bottomrule
% \end{tabular}
% \caption{\label{tab:additional_methods_performance}different MLLM's align result on Geometry3K}
% \end{table}

% \subsubsection{The effect of align module in LLM}

\begin{table}
\centering
\begin{tabular}{cccc}
\toprule
 Predictor & Completion & Choice & Steps \\
\midrule
Traversal       & 68.9 & 76.6 & 2.85 - 6.03 \\
Claude 3.5 Sonnet  & 70.2 & 77.5 & 2.75 - 4.60  \\
GPT-4o       & 69.4 & 77.2 & 2.62 - 4.41 \\
Gemini 2 & 69.8 & 77.2 & 2.52 - 4.29 \\
o3-mini  & \textbf{70.6} & \textbf{77.8} & \textbf{2.31 - 4.12} \\
\bottomrule
\end{tabular}
\caption{Illustrating the effect of different LLMs used in theorem predicton on Geometry3K. All evaluated LLMs achieve comparable performance. }
\label{tab:theorem_results} 
\end{table}

\begin{figure}[t]
  \centering
  \includegraphics[width=\columnwidth]{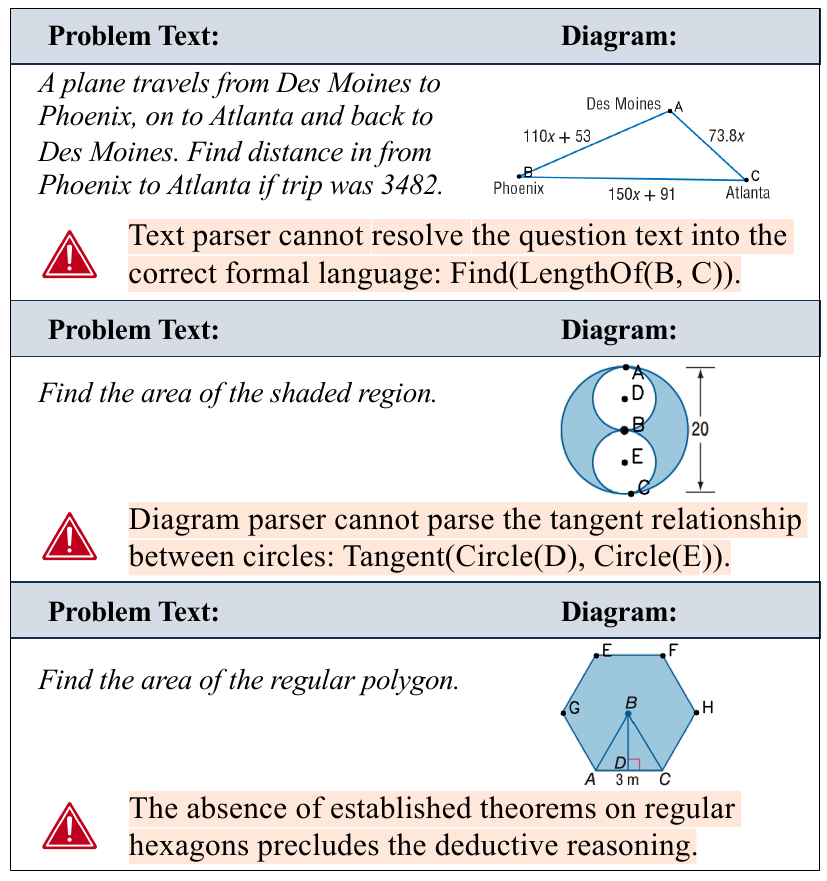}
  \caption{Illustrating the limitations in current GPS framework.}
  \label{fig:limitations}
\end{figure}

\section{Conclusion}
In this work, we present Pi-GPS, a novel framework that integrates diagrammatic information to enhance geometry problem solving by resolving textual ambiguities. Central to our approach is a rectifier-verifier module, where the rectifier leverages MLLMs to refine textual descriptions using diagrammatic context, while the verifier ensures geometric consistency. This framework significantly improves problem representation, thereby improving the problem solving performance.
Empirical evaluations on the Geometry3K and PGPS9K benchmarks demonstrate that Pi-GPS outperforms state-of-the-art neural-symbolic methods, achieving nearly a 10\% performance gain on Geometry3K. These results advocate the critical role of ambiguity resolution in multimodal mathematical reasoning, a challenge that has been largely overlooked and warrants greater attention from the research community.
% We hope this work inspires further research into integrating multimodal reasoning techniques for improving AI-driven geometry problem solving.

% In this paper, we introduce Sync-GPS, a novel symbolic geometry problem-solving framework enhanced with a comprehensive alignment module leveraging MLLMs. By integrating rule-based text parsing, advanced diagram parsing with PGDPNet, and our proprietary Sync module, Sync-GPS ensures precise alignment between textual and visual information. Experiments on the Geometry3K and PGPS9K datasets demonstrate that Sync-GPS outperforms existing symbolic methods in accuracy and interpretability while maintaining competitive performance against neural-based models with greater data efficiency. Additionally, our alignment framework effectively mitigates MLLM hallucinations, enhancing the reliability of solutions. Future work will focus on refining parsers, expanding the theorem library, and exploring applications in broader multimodal domains to further improve problem-solving capabilities.

\section{Limitations}
While this work successfully identifies text ambiguity and introduces a dedicated module to resolve it, significantly enhancing system performance. Several limitations still remain as illustrated in Figure~\ref{fig:limitations}. These limitations highlight key areas for future improvement and offer directions for advancing automated geometric proble solving systems.

\begin{itemize}
    \item \textbf{Limited Text Parsing Capability}: The current text parser struggles to accurately map certain syntactic variations to their formalized representations. Despite the integration of our text disambiguation module, these challenges persist, often leading to incomplete or erroneous formalizations. 
    % Consequently, the accuracy of subsequent geometric problem solving processes is compromised.

    \item \textbf{Inadequate Diagram Parsing for Complex Relations}: The diagram parser struggles to accurately identify complex geometric relationships, such as tangency, due to their subtle and often implicit nature. This limitation hampers precise geometric analysis and interpretation, as misrecognition can distort structural understanding and compromise downstream computations.
    
    \item \textbf{Insufficient Theorem Base}: The absence of essential theorems necessary for solving specific problem categories significantly constrains the system's ability to generate comprehensive and accurate solutions. For example, in the case of a regular hexagon, the fundamental theorem asserting that each interior angle measures 120 degrees is critical for various geometric deductions.
\end{itemize}

{
    \small
    \bibliographystyle{ieeenat_fullname}
    \bibliography{main}
}

\end{document}